\documentclass[letterpaper, 10 pt, conference]{ieeeconf}  

\newtheorem{mydef}{Definition}
\IEEEoverridecommandlockouts                              

\usepackage{graphics} 
\usepackage{epsfig} 
\usepackage{amsmath} 
\usepackage{amssymb}  
\usepackage{multirow}
\usepackage{cite}
\makeatletter
\let\NAT@parse\undefined
\makeatother
\usepackage{hyperref}

\title{\LARGE \bf
Generalizing to New Domains by Mapping Natural Language  to Lifted LTL
}

\author{Eric Hsiung, Hiloni Mehta, Junchi Chu, Xinyu Liu, Roma Patel, Stefanie Tellex, George Konidaris
\thanks{The authors are with the Brown University Department  of  Computer Science, 115 Waterman Street, Providence, RI 02912. Email: \{eric\_hsiung, hiloni\_mehta, junchi\_chu, xinyu\_liu, roma\_patel1, stefanie\_tellex, george\_konidaris\}@brown.edu}
}

\begin{document}

\maketitle
\thispagestyle{empty}
\pagestyle{empty}

\begin{abstract}Recent work on using natural language to specify commands to robots has grounded that language to LTL. However, mapping natural language task specifications to LTL task specifications using language models require probability distributions over finite vocabulary. Existing state-of-the-art methods have extended this finite vocabulary to include unseen terms from the input sequence to improve output generalization. However, novel out-of-vocabulary atomic propositions cannot be generated using these methods. To overcome this, we introduce an intermediate \textit{contextual query} representation which can be learned from single positive task specification examples, associating a contextual query with an LTL template. We demonstrate that this intermediate representation allows for generalization over unseen object references, assuming accurate groundings are available. We compare our method of mapping natural language task specifications to  intermediate contextual queries against state-of-the-art CopyNet models capable of translating natural language to LTL, by evaluating whether correct LTL for manipulation and navigation task specifications can be output, and show that our method outperforms the CopyNet model on unseen object references. We demonstrate that the grounded LTL our method outputs can be used for planning in a simulated OO-MDP environment. Finally, we discuss some common failure modes encountered when translating natural language task specifications to grounded LTL.

\end{abstract}

\section{INTRODUCTION}

Programming robots to accomplish tasks often requires human experts to design robot controllers.
Communicating task specifications to a robot in natural language would be preferable because non-expert humans could specify tasks to robots as if speaking to another human.
However, in order for a robot to accomplish a task, the natural language task specification must be grounded into a form the robot can interpret. Recent work has focused on grounding natural language commands to Linear Temporal Logic, or LTL\cite{pnueli1977}, which can be used as a reward function for a planner\cite{sadigh2014,FuT14,OhPNHPT19,Camacho2019LTLAB}. 

Current state of the art methods \cite{BergBMRPT20, GopalanAWT18, OhPNHPT19, Danas2019FormalDM} use Seq2Seq \cite{SutskeverVL14} models to learn to ground natural language into LTL, given training data that pairs natural language commands and corresponding LTL formulae. These approaches have led to substantial progress, but have two major drawbacks.
First, Seq2Seq models are trained on finite vocabularies that are present either in the training set or have been determined a priori. The output of these models can only contain elements drawn from their vocabulary. Some techniques\cite{GuLLL16} can extend the vocabulary to include out-of-vocabulary (OOV) words from the input sequence, effectively allowing unseen words from the input sequence to be incorporated directly into the output sequence, but these techniques cannot generate words outside the extended vocabulary. As a consequence, words outside the vocabulary or input sequence cannot be generated.
Second, these models always output grounded LTL, so the learned models are linked to the specific domains they were trained on. It would be desirable for lifted LTL to be generated so that LTL task structures can be transferred between environmental domains.

We address these two issues by introducing an intermediate representation that links a \textit{contextual query} representation with a templated LTL task representation, and we contribute a method for associating the two representations to generalize over unseen objects. As a result, LTL can be generalized over objects by evaluating templated LTL on unseen objects. We apply our method on simulated manipulation and navigation domains to demonstrate improved generalization over unseen objects compared with state-of-the-art natural language to LTL Seq2Seq models.
\section{BACKGROUND}
In this work, we represent the environments in which a task is executed using object-oriented MDP formalism, and utilize LTL to represent temporal tasks. Grounding is the association of symbols with objects or states in the environment, and is used to connect language and atomic propositions to objects and states in an OO-MDP.
\subsection{Object-Oriented MDPs (OO-MDPs)}
A Markov Decision Processs (MDP) is a tuple of states $S$, actions $A$, transition function $T$ capturing environment dynamics, reward model $R$, and discount factor $\gamma$ represented as $(S,A,T,R,\gamma)$, which can be used to model sequential decision making in a environment. In an OO-MDP\cite{Diuk2008AnOR}, the MDP definition is extended to include \textit{object classes} which are defined by sets of attributes, and \textit{propositional functions} which operate on specific object states to return Boolean values, often comparing the input object attributes. Propositional functions are well-suited for generating atomic propositions about the environment, and form a basis over which lifted LTL can be expressed.
\subsection{Linear Temporal Logic (LTL)}
LTL is a modal temporal logic that follows a particular grammatical syntax: $\phi::=\pi \:|\: \neg\phi \:|\: \phi \wedge \psi \:|\: \phi \vee \psi \:|\: \mathcal{G}\phi \:|\: \mathcal{F}\phi \:|\: \phi \mathcal{U}\psi$, where $\phi$ is the task specification, $\phi$ and $\psi$ are LTL formulae; atomic proposition $\pi$ is drawn from a set $\Pi$ of possible propositions;  $\mathcal{F}$, $\mathcal{G}$,  $\mathcal{U}$ denote the \textit{finally}, \textit{globally} or \textit{always}, and \textit{until} temporal operators; and $\neg$, $\wedge$, $\vee$ represent logical operators \textit{negation}, \textit{and}, and \textit{or}. LTL can be used to represent temporal tasks, and reflects the desired temporal evolution of a set of atomic propositions. Furthermore, a LTL formula can be represented as a B\"{u}chi automaton\cite{Bchi1990OnAD}. Pairing the automaton states with MDP states results in a product state which can be used to define a product MDP, over which planning can be done once the product MDP is solved.
\subsection{Grounding and Labeling Functions}
In the context of an OO-MDP, a grounding function \cite{Vogt2006LanguageEA} maps natural language nouns and adjectives to objects and their attribute values in the environment, whereas a labeling function can be considered the inverse of grounding: mapping states or objects to a set of symbols. Atomic propositions are associated with states, so a labeling function $L: S\rightarrow 2^{\Pi}$ maps states to the Boolean values for the set of atomic propositions $\Pi$ under consideration.
\begin{figure*}[]
    \centering
    \small Training Pipeline (a)\\[\smallskipamount]
    \includegraphics[width=\textwidth]{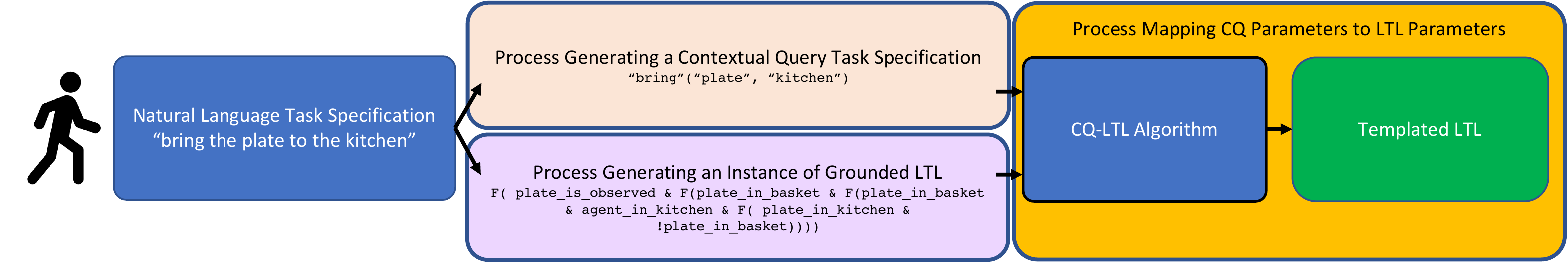}
    \\[\smallskipamount]
    \small Evaluation Pipeline (b)\\[\smallskipamount]
    \includegraphics[width=\textwidth]{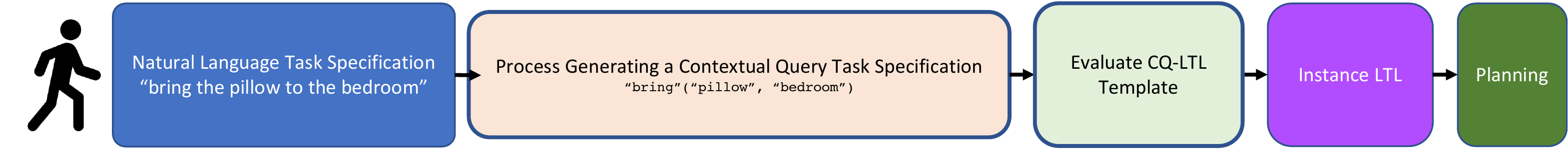}
    \\[\smallskipamount]
    \small High Level Example (c)\\[\smallskipamount]
    \includegraphics[width=\textwidth]{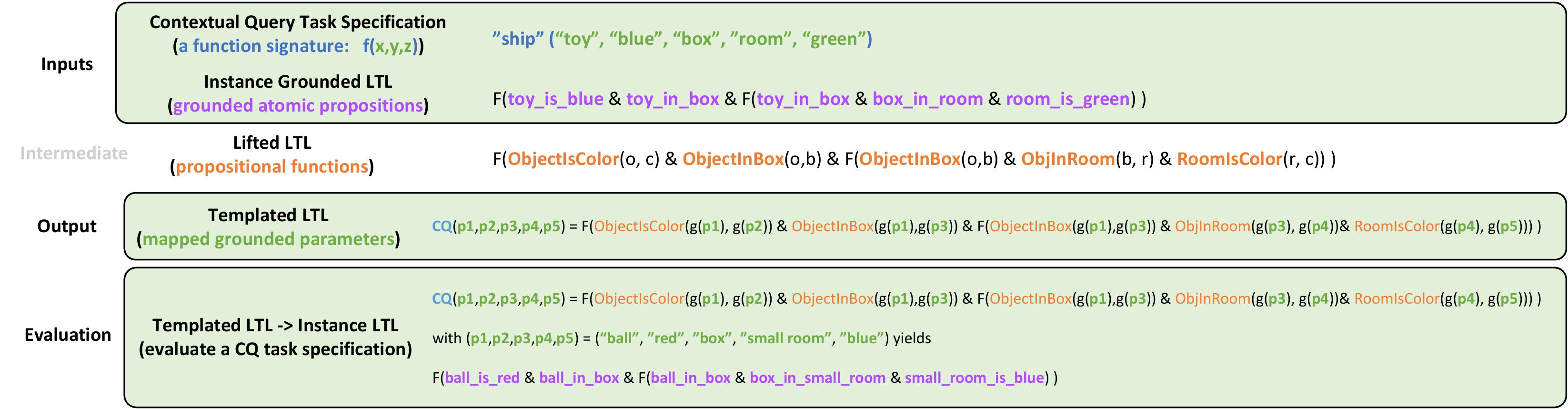}
    \caption{The training and evaluation pipelines for CQ-LTL. (a) The training pipeline incorporates any two processes which can generate a \textit{contextual query task specification} and a corresponding instance of \textit{grounded LTL task specification} from a natural language task specification. These serve as the inputs to the CQ-LTL algorithm which outputs \textit{templated LTL}. (b) The evaluation pipeline only requires the process for generating \textit{contextual query task specifications} from a natural language task specification, and outputs \textit{grounded LTL} by evaluating the templated LTL with the contextual query task specification. The instance of grounded LTL can subsequently be used for planning. (c) High level example indicating the transition from an instance of grounded LTL to lifted LTL, and then association of contextual query parameters with propositional function parameters in the lifted LTL to achieve templated LTL. Finally, for evaluation, the templated LTL is evaluated according to a contextual query task specification to generate grounded LTL.}
    \label{train-eval-diagram}
\end{figure*}
\section{RELATED WORK}
Prior work \cite{GopalanAWT18,OhPNHPT19,BergBMRPT20} uses supervised learning to train Seq2Seq models to output grounded LTL task specifications from natural language. Gopalan et al. \cite{GopalanAWT18} discussed challenges relating to such language model generalization, and introduced and demonstrated the ability for Seq2Seq models to ground natural language to geometric LTL, utilizing the framework of grounding natural language to reward functions introduced by MacGlashan et al. \cite{MacGlashanBdLMS15}, where reward functions task specifications were expressed as conjunctions of propositional functions learned from demonstration. Berg et al. \cite{BergBMRPT20} focused on the generalization problem of grounding \textit{unseen} language to LTL by applying CopyNet \cite{GuLLL16}, a Seq2Seq model with a copying mechanism, to copy out-of-vocabulary words present in the input command to the output LTL. The contribution improved LTL generation for cases where the atomic propositions could be directly represented by the natural language vocabulary. Oh et al. \cite{OhPNHPT19}  introduced more efficient planning methods for solving LTL task specifications in MDPs using product state abstractions, and demonstrated planning on navigation LTL task specifications generated from Seq2Seq models.

Other work \cite{Patel2020GroundingLT,WangTLHL20,Danas2019FormalDM} has also considered grounding natural language to LTL task specifications with improved accuracy using example trajectories and human-feedback, using formal verification and optimization methods. Patel et al. \cite{Patel2020GroundingLT} introduced a semi-supervised method for grounding language to LTL via generative models and formal verification, considering whether candidate LTL satisfies trajectories. Wang et al. \cite{WangTLHL20}  introduced an algorithm for learning LTL task specifications of arbitrary complexity from human-feedback by a process of eliminating sub-optimal trajectories. Danas et al. \cite{Danas2019FormalDM} contributed a three step process for recovering from language grounding errors by performing beam search within a Seq2Seq model to determine the top most likely LTL formulae, differentiating trajectories via maximal semantic differencing, and finally requesting human feedback to clarify which trajectories match the desired task specification.

A common theme all these works share is their focus on the domain of grounding \textit{navigation} natural language task specifications to LTL task specifications, where the atomic propositions can easily be drawn from existing natural language vocabulary as in the case of Berg et al.\cite{BergBMRPT20}. The exception to this was MacGlashan et al.\cite{MacGlashanBdLMS15}, where the focus was on representing atemporal task specifications. Our work builds off of the work of Gopalan et al. \cite{GopalanAWT18}, MacGlashan et al. \cite{MacGlashanBdLMS15}, and Berg et al. \cite{BergBMRPT20} in that we employ propositional functions to generate atomic propositions. Our contributions differ from prior work in that we consider lifted LTL representations in order to generalize task specifications over objects, and consider these representations for both manipulation and navigation tasks.
\section{APPROACH}
Current state of the art Seq2Seq models that translate directly from natural language to LTL typically must account for atomic propositions as part of the LTL output sequence vocabulary \cite{GopalanAWT18}. For navigation tasks, nouns  are typically grounded to locations in the environment, so input nouns can be directly used to represent the atomic propositions in the output. The natural language command \textit{go to CVS and then go to the park} would map to \textit{$\mathcal{F}$(CVS $\wedge$ $\mathcal{F}$(park))}, where atomic propositions \textit{CVS} and \textit{park}  imply that the agent is at the CVS location and park location respectively. For more complex tasks, such as object manipulation, individual nouns do not directly map to atomic propositions in the output. Rather, combinations of nouns and their attributes correspond to atomic propositions: \textit{make sure the light is on before putting the ball in the small bucket} would correspond with \textit{$\mathcal{F}$(light\_on $\wedge$ $\mathcal{F}$(ball\_in\_small\_bucket))}, where the atomic proposition \textit{ball\_in\_small\_bucket} is dependent on attributes of the ball and the small bucket.

The previous examples show that standard Seq2Seq models require compound atomic propositions to be part of the output vocabulary. Since an environment could have an arbitrary number of objects with arbitrary attributes,  the output vocabulary grows exponentially with the number of objects and attributes. For Seq2Seq models to generalize across objects, the output vocabulary must always be expanded to ensure the target objects and attributes are part of the vocabulary. In practice, this is impractical, since Seq2Seq models must first be trained on a dataset using finite vocabulary. In other words, a Seq2Seq model is trained on a specific distribution of objects which correspond with its output vocabulary, and these models can fail to generalize on unknown objects and attributes outside the vocabulary. CopyNet alleviates the issue of unseen words by copying words from the input to the output, but this requires atomic propositions to be drawn from existing natural language vocabulary.

To resolve generalizing over objects, we propose learning \textit{templated LTL} to represent parameterized \textit{task classes} via one-shot learning from example task specifications. \textit{Task specifications} couple task parameters with a task description, and can be represented in multiple ways.  Our approach uses contextual queries as an intermediate, functional representation of the set of task specifications belonging to a task class. A \textit{contextual query} is a human interpretable function signature $f(x,y,z,...)$ that is representative of a task specification, where $f$ is a human interpretable description corresponding with a task class, such as ``pick up objects'' and $(x,y,z,...)$ are human interpretable ordered parameters of the task class. Since contextual queries represent task classes, we can associate templated LTL with a contextual query, much like a function body can be defined for a function signature. If natural language object references can be directly mapped to the relevant parameter positions in a contextual query associated with templated LTL, then evaluating the contextual query results in an instance of \textit{grounded LTL}, corresponding to the task specification.

We hypothesize that associating a contextual query with templated LTL leads to better task generalization across objects, and that accuracy improves with grounding accuracy. Specifically, we hypothesize that Seq2Seq models will succeed on navigation tasks, and will not be performant on manipulation tasks due to atomic propositions missing from the output vocabulary. Contextual queries associated with templated LTL alleviate this issue by accepting out-of-vocabulary objects and outputting the corresponding out-of-vocabulary atomic propositions. Thus, we expect contextual queries with templated LTL to succeed on both navigation and manipulation task specifications.

\subsection{Task Specifications and Task Classes}
A single task can be specified using various equivalent representations. A task specification is way to describe or represent a particular task. A task class is a collection of tasks which can be grouped together according to some property.
\mydef{A \textit{task specification} $\tau = (T, P, \mathcal{R})$ is a tuple of a task descriptor $T$, set of values $P$ that specify the parameterized task, and representation operator $\mathcal{R}$ that operates on $(T,P)$ to output a representation of the task specification.}
\mydef{A \textit{task class} $\mathcal{T}$ is a set of all task specifications $\tau$ that share equivalent task descriptors $T$. Different representations of the same task specification belong to the same task class.}

Natural language is frequently used to specify tasks: \textit{open the refrigerator} and \textit{open the garage} are both task specifications which belong to the same task class, whereas \textit{pick up a red apple} is a task specification that belongs to a different task class. A task specified in natural language has an equivalent representation in LTL, as well as an equivalent representation as a contextual query.

\subsection{Contextual Queries}
\mydef{A \textit{contextual query} $C$ is the functional representation of a task class $\mathcal{T}=\{(T_i, P_i, \mathcal{R}): \forall i, T_i = T, |P_i|=N\}$ for a constant $N\in \mathbb{N}$ and task descriptor $T$. Let $\mathcal{T}_F$ be the subset of $\mathcal{T}$ where representation $\mathcal{R}$ is a functional representation $F$. Let $\mathcal{T}_{\mathcal{R}'}$ be the subset of $\mathcal{T}$ where representation $\mathcal{R}$ is the representation $\mathcal{R'}$. Then $C$ also is a representation transform $C : \mathcal{T}_F \rightarrow \mathcal{T}_{\mathcal{R}'}$, such that $C(T, P, F) = T(^*P)=(T,P,\mathcal{R'})$, where $T: (\cup_i P_i)^{N} \rightarrow \mathcal{T}_{\mathcal{R}'}$, and $^*P$ is a $N-$tuple that represents a specific ordering of the elements of $P$.}

In our model, we introduce CQ-LTL, which is a type of contextual query that transforms functional task specifications to LTL task specifications. The task descriptor $T$ is a human-specified function signature $T(\cdot)$, which is a human interpretable description of the task class, and can be considered to be a label for the task class. The parameters of the function are ordered, and correspond to the parameters of the task class.
Because contextual queries are a representation of a task class, equivalent representations can be associated with the task class to complete the mapping. In our approach, we associate lifted LTL with the contextual query, where lifted LTL is the alternative representation of the task class.

\subsection{Lifted LTL and Templated LTL}
Our approach takes in instances of \textit{grounded LTL}, which are examples of grounded LTL task specifications consisting of atomic propositions that ground to objects in the environment. Given an instance of grounded LTL and a set of propositional functions, we can derive \textit{lifted LTL}, which is LTL consisting of propositional functions. Propositional functions themselves are functions of objects, so lifted LTL is not resolvable until arguments are provided, as shown in Fig. \ref{train-eval-diagram}.

In order to provide values for the parameters in the lifted LTL, we associate parameters in the contextual query with parameters in the lifted LTL. This parameter association allows lifted LTL to become \textit{templated LTL}. Thus, when the corresponding contextual query is evaluated with a new set of arguments, these values are propagated according to the mapping to the parameters in the propositional functions. Evaluating the propositional functions on objects generates atomic propositions. Thus \textit{templated LTL} can be used to generate \textit{grounded LTL} via contextual query evaluation.

\subsection{CQ-LTL Algorithm}
We present a simple algorithm to derive \textit{templated LTL} from a pair of equivalent task specifications. The algorithm takes two inputs: (1) a \textit{contextual query} task specification, and (2) the corresponding instance of \textit{grounded LTL} task specification. We assume that the environment is represented by an underlying OO-MDP that has a set of propositional functions defined over the set of objects in the OO-MDP. We also assume a grounding function is available that maps from the domain of the contextual query to the objects of the OO-MDP.

The algorithm has 2 steps. First, the instance of \textit{grounded LTL} is converted to \textit{lifted LTL} by substituting the atomic propositions for the corresponding propositional functions. Second, the mapping from contextual query parameters to propositional function parameters is determined by matching the groundings of the contextual query task specification parameters to the propositional function parameters. The process is illustrated in Fig. \ref{train-eval-diagram}. To obtain valid parameter mappings, the grounded inputs must be distinct from one another, otherwise it is possible for two or more contextual query parameters to map to the same propositional function parameter, resulting in an invalid parameter mapping.

\section{Experiments}
We implemented various pipelines to demonstrate the utility of templated LTL, evaluating how well it generalizes on seen and unseen objects, and compared it against the CopyNet model that was part of our pipeline. We then provided a demonstration of the evaluation pipeline from natural language to planning under different environments to illustrate how templated LTL can generalize between domains across objects.

\subsection{Experimental Domains}
We demonstrated planning on navigation tasks and object manipulation tasks in environments implemented as OO-MDPs domains. The \textit{Toy} domain is a taxi-like discrete state-space domain that supports both object manipulation and navigation tasks, and represents a robot gripper moving around to pickup objects of different shapes and colors, placing them into a box which can be re-positioned by the gripper. Grid cells in the environment are represented by squares with a color attribute. The size of the state space in the \textit{Toy} environment is upper bounded by $N(N+1)(N+2)^k$, where $N$ is the number of grid cells, and $k$ is the number of toys in the environment.

\subsubsection{Navigation} We considered three navigation LTL task structures representing unconstrained waypoint navigation: 
$$\begin{array}{ccccc}
    \mathcal{F} (\phi)  &|& \mathcal{F} (\phi \wedge \mathcal{F} (\psi)) &|& \mathcal{F} (\phi\wedge \mathcal{F} (\psi\wedge \mathcal{F} (\varphi))).
\end{array}$$ $\phi$, $\psi$, and $\varphi$ were drawn from atomic propositions representing the agent at particular locations, using the existing natural language vocabulary for locations.

\subsubsection{Manipulation} We considered four manipulation LTL task structures---\textit{move to} (task for robot to move to the mentioned object), \textit{pickup} (task to pickup the mentioned object), \textit{pickup colored} (task to pickup an object with mentioned attributes), and \textit{ship} (task for putting the mentioned object in a container and moving the container to the desired location).
For \textit{move to} and \textit{pickup}, the $\mathcal{F} (\phi)$ LTL representation was used, with $\phi$ representing whether the agent is at the object location in the \textit{move to} case, and whether the agent holds the object in the \textit{pickup} case. For \textit{pickup colored}, $\mathcal{F}(\phi \wedge \psi)$ was used, with $\phi$ representing the agent holding the object, and $\psi$ representing if the object possessed a matching attribute value. Finally, for \textit{ship}, $\mathcal{F}(\phi \wedge \mathcal{F}(\phi \wedge \psi))$ was used, with $\phi$ representing if the object was in a specific container, and $\psi$ representing if the container was in the desired location.
\begin{table}[]
    \centering
    \caption{Dataset, training, and evaluation details}
    \label{dataset-details}
\begin{tabular}{ |c|l|l|  }
 \hline
 & Manipulation & Navigation \\
 \hline
 \hline
 \textit{Seen}, $S$ & $|S|=4650$, $|V_S|=360$ & $|S|=8007$, $|V_S|=42$\\
 \textit{Unseen}, $U$ & $|U|=1764$, $|V_U|=340$ &  $|U|=378$, $|V_U|=23$\\\hline
 Training & $T\subset S, |T|=1920$ & $T\subset S, |T|=840$\\
 Validation & 10\% of $T$ & 10\% of $T$\\
 \hline
 Task& \textit{move\_to} (150, 36) & \textit{navigate\_one} (51, 18) \\
 Classes & \textit{pickup} (450, 108)&  \textit{navigate\_two} (612, 72) \\
 ($|\mathcal{T}_S|$, $|\mathcal{T}_U|$) & \textit{pickup\_colored} (2700, 648) & \textit{navigate\_three} (7344, 288) \\
  & \textit{ship} (1350, 972) &  \\
 \hline
\end{tabular}
\end{table}
\begin{figure}[]
    \centering
    \includegraphics[width=\columnwidth]{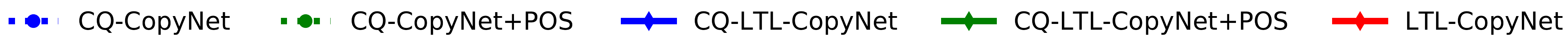}\\
    \includegraphics[width=0.5\columnwidth]{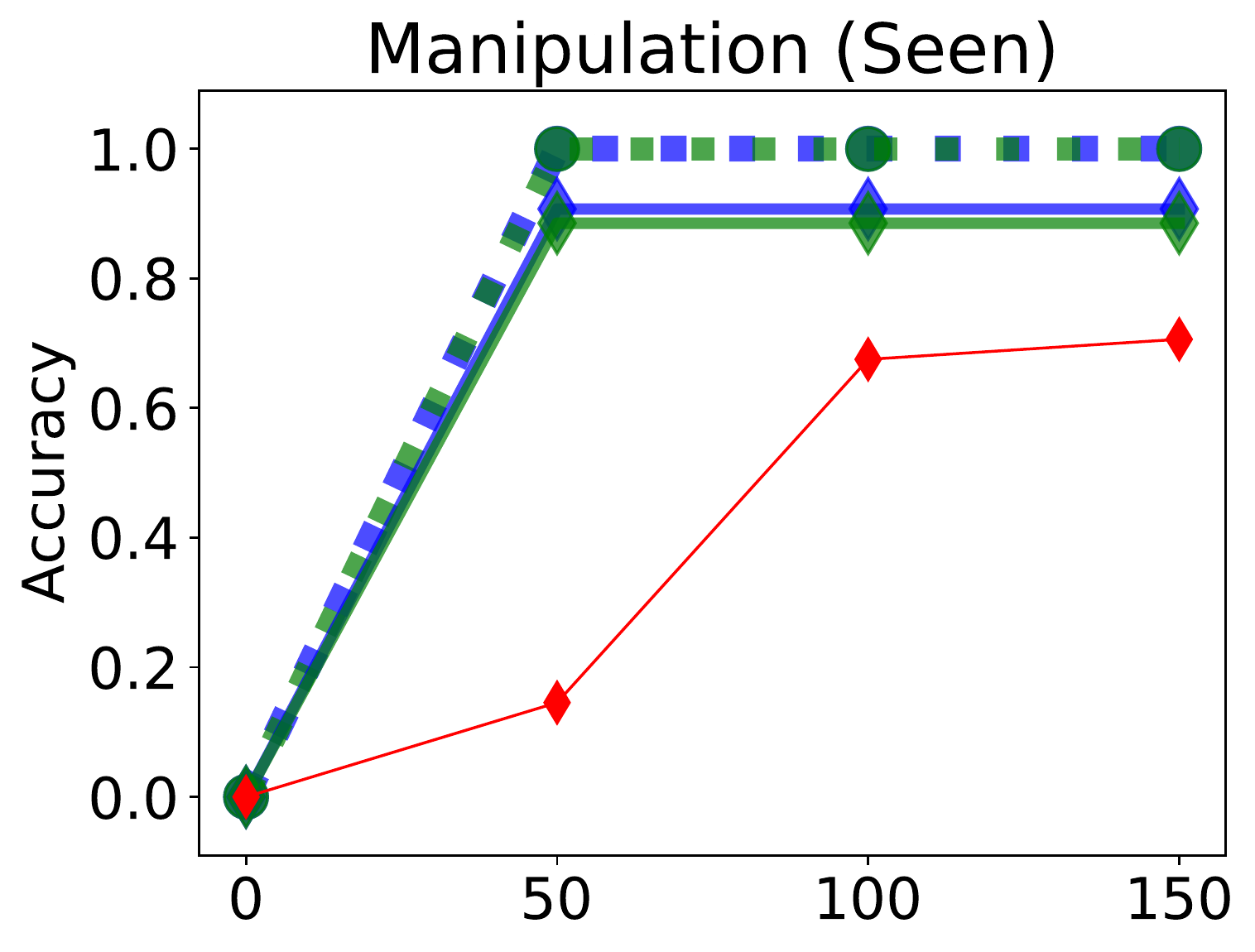}\hfill
    \includegraphics[width=0.5\columnwidth]{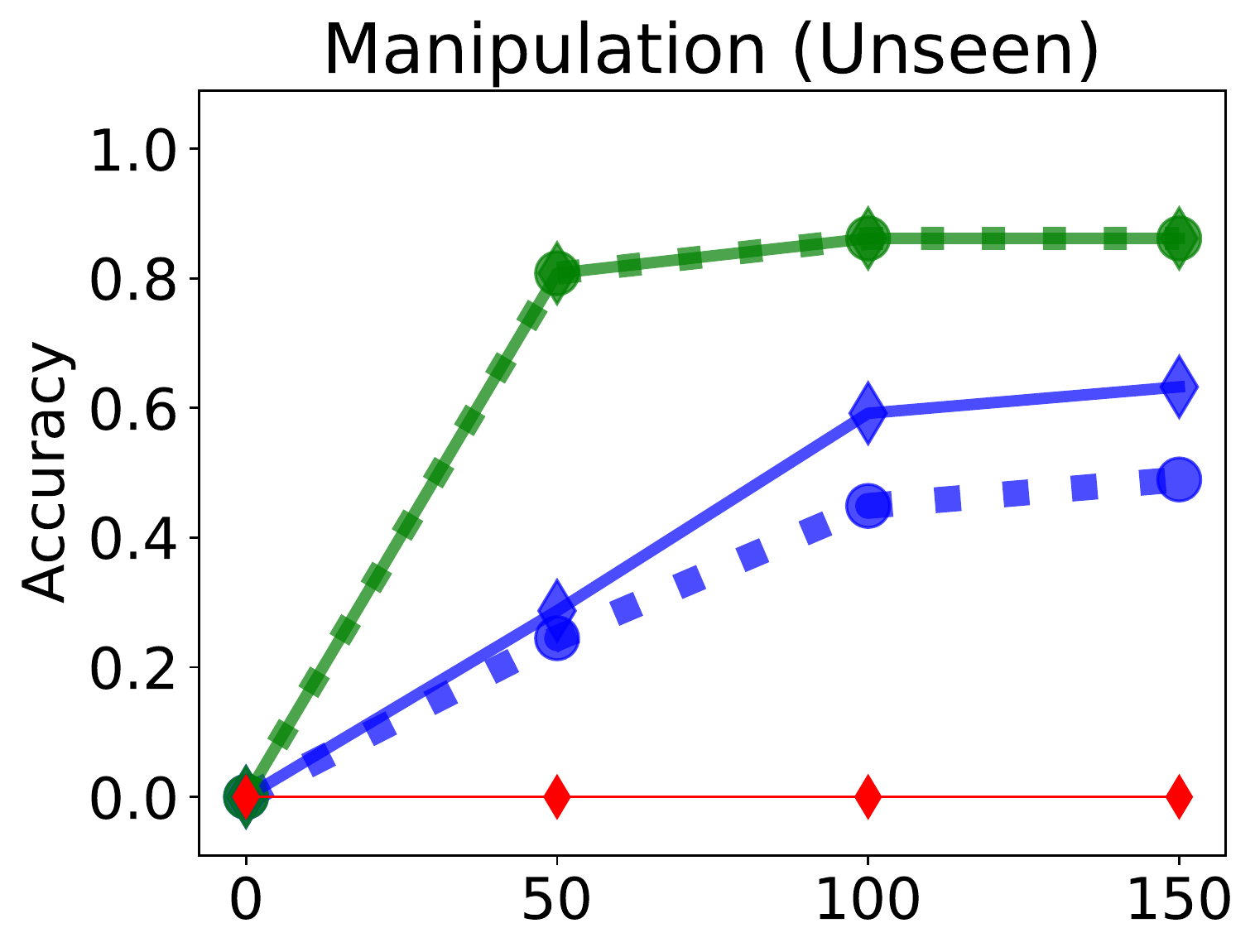}
    \\
    \includegraphics[width=0.5\columnwidth]{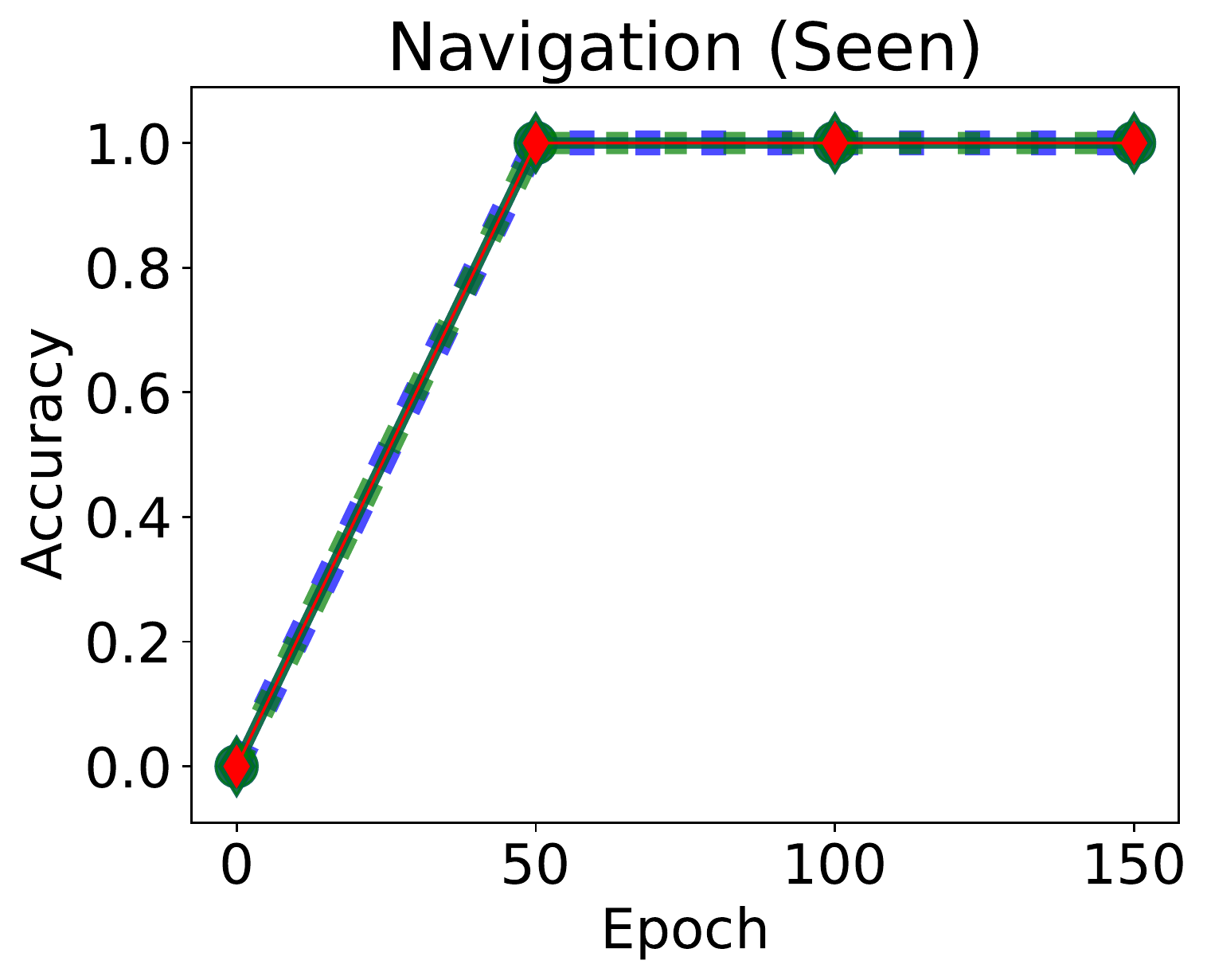}\hfill
    \includegraphics[width=0.5\columnwidth]{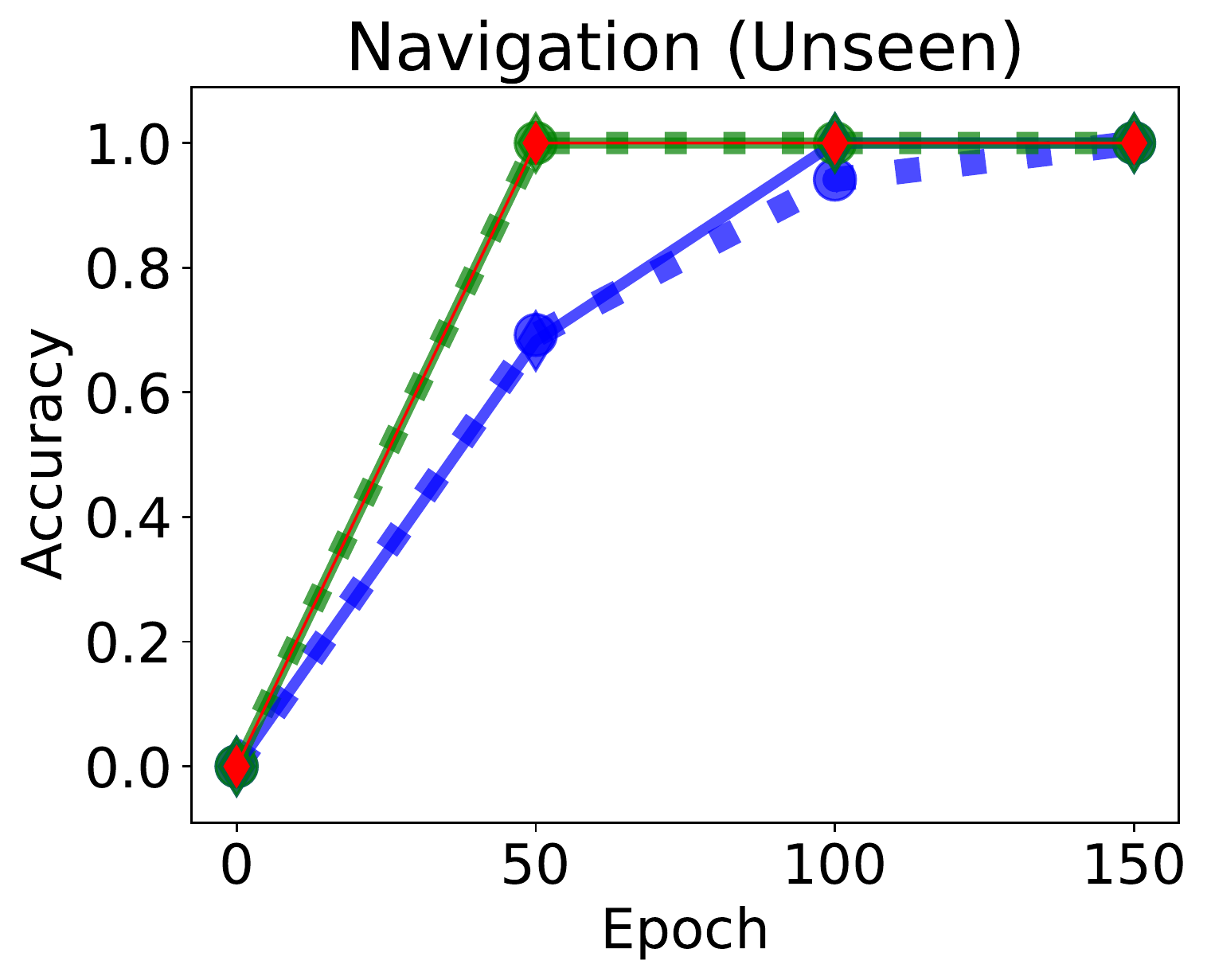}
    \\
    \caption{Overall language model task specification accuracy for manipulation and navigation tasks using seen and unseen vocabulary. \textbf{Top Row}: Manipulation task accuracy. \textbf{Bottom Row}: Navigation task accuracy. \textbf{Left Column}: Tasks specified with seen vocabulary. \textbf{Right Column}: Tasks specified with unseen vocabulary. \textbf{Legend:} Models outputting contextual queries are dotted with circles, and models outputting LTL are solid with diamonds.}
    \label{overall-accuracy}
\end{figure}
\begin{figure*}[]
    \centering
    \includegraphics[width=\textwidth]{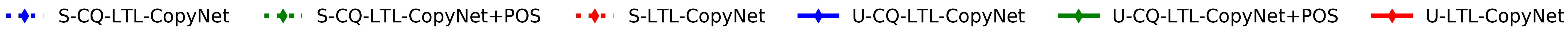}\\
    \includegraphics[width=0.1428\textwidth]{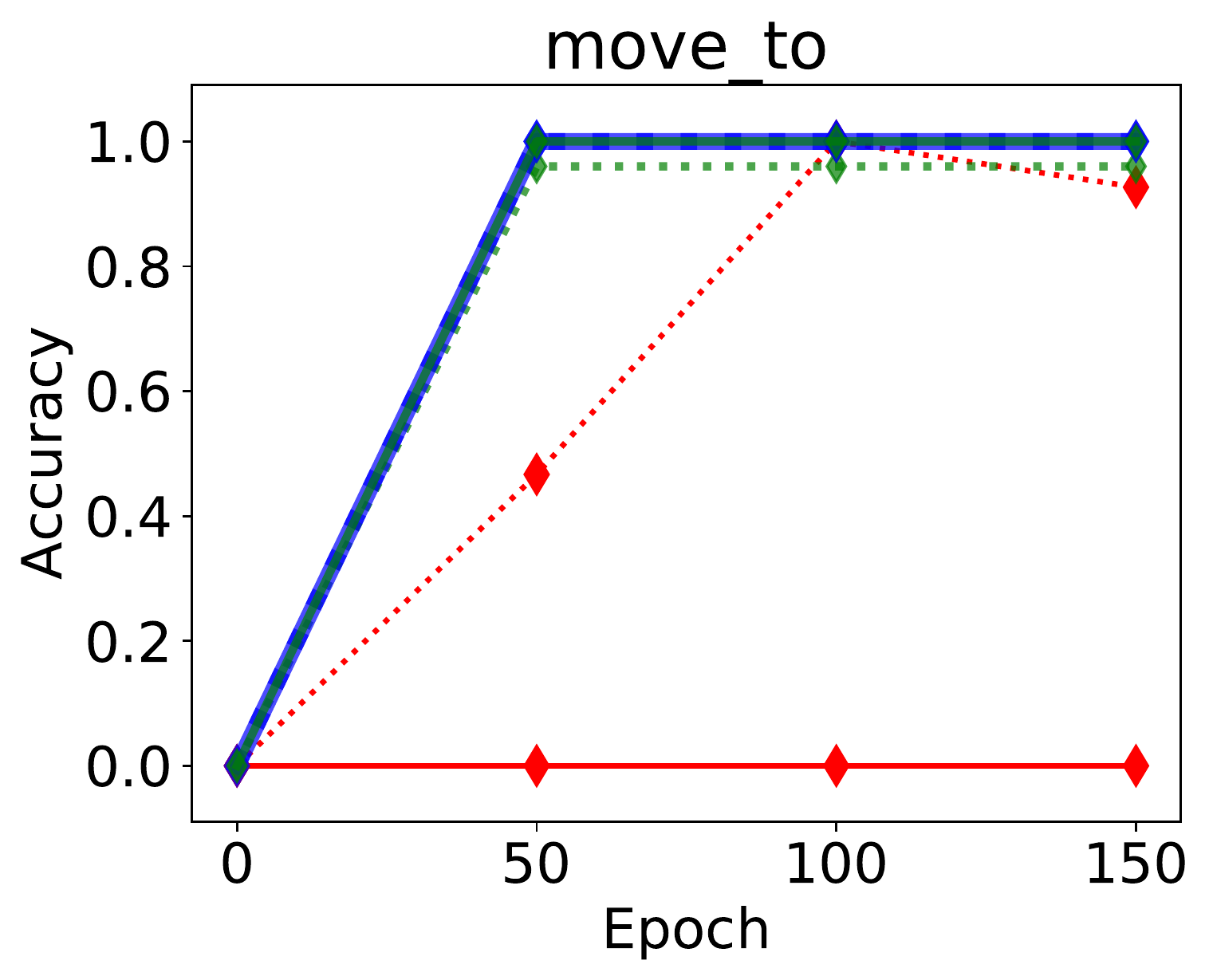}\hfill
    \includegraphics[width=0.1428\textwidth]{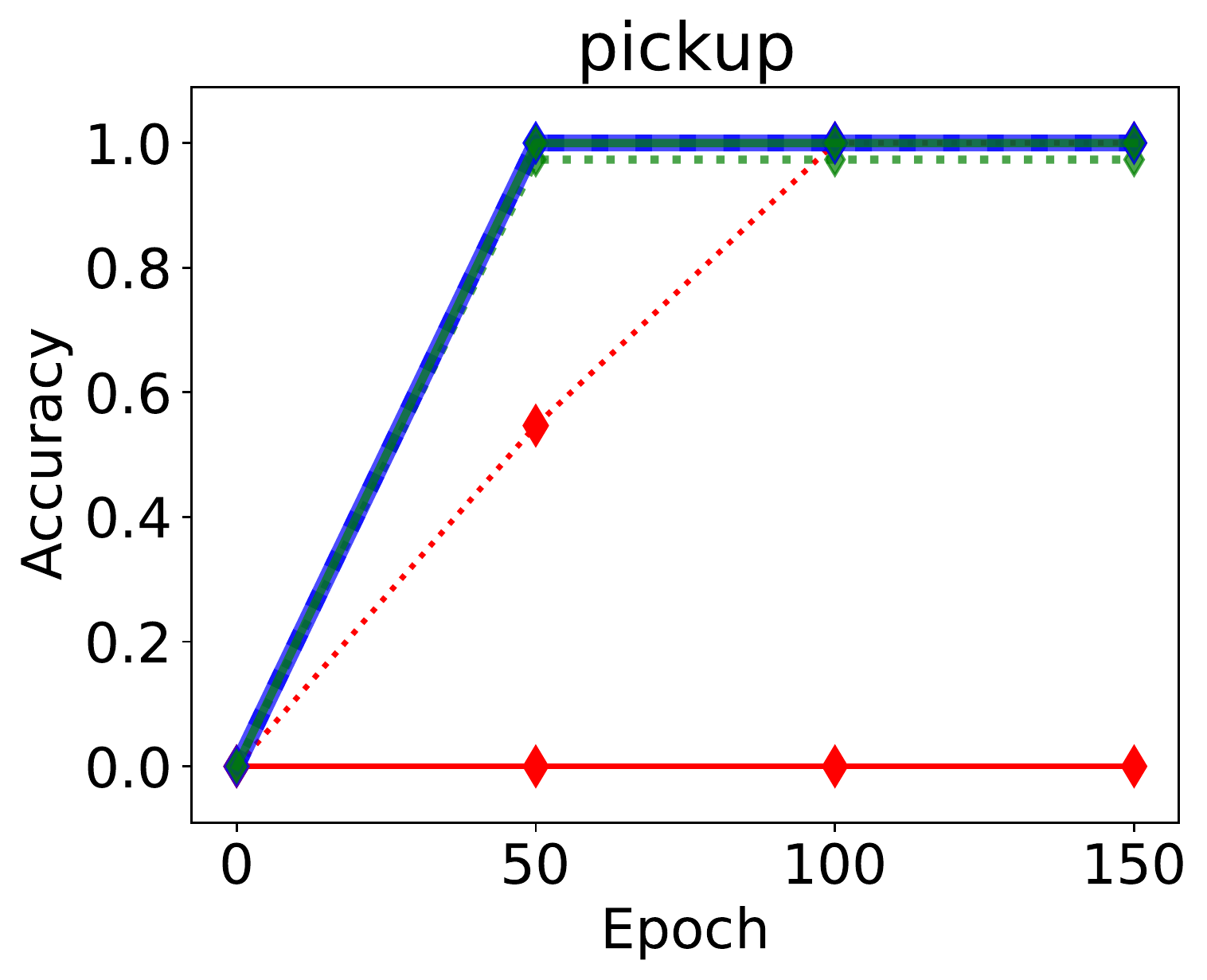}\hfill
    \includegraphics[width=0.1428\textwidth]{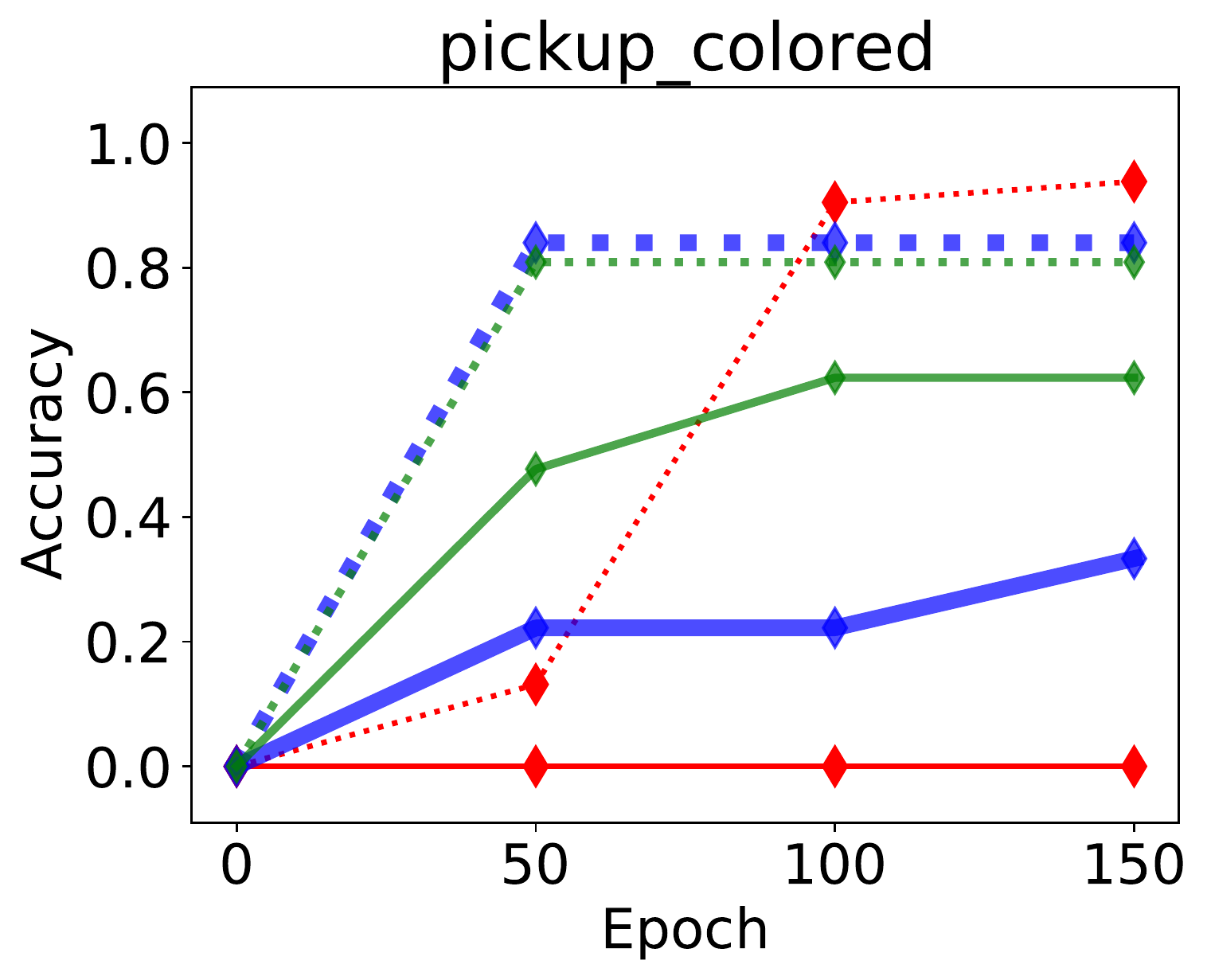}\hfill
    \includegraphics[width=0.1428\textwidth]{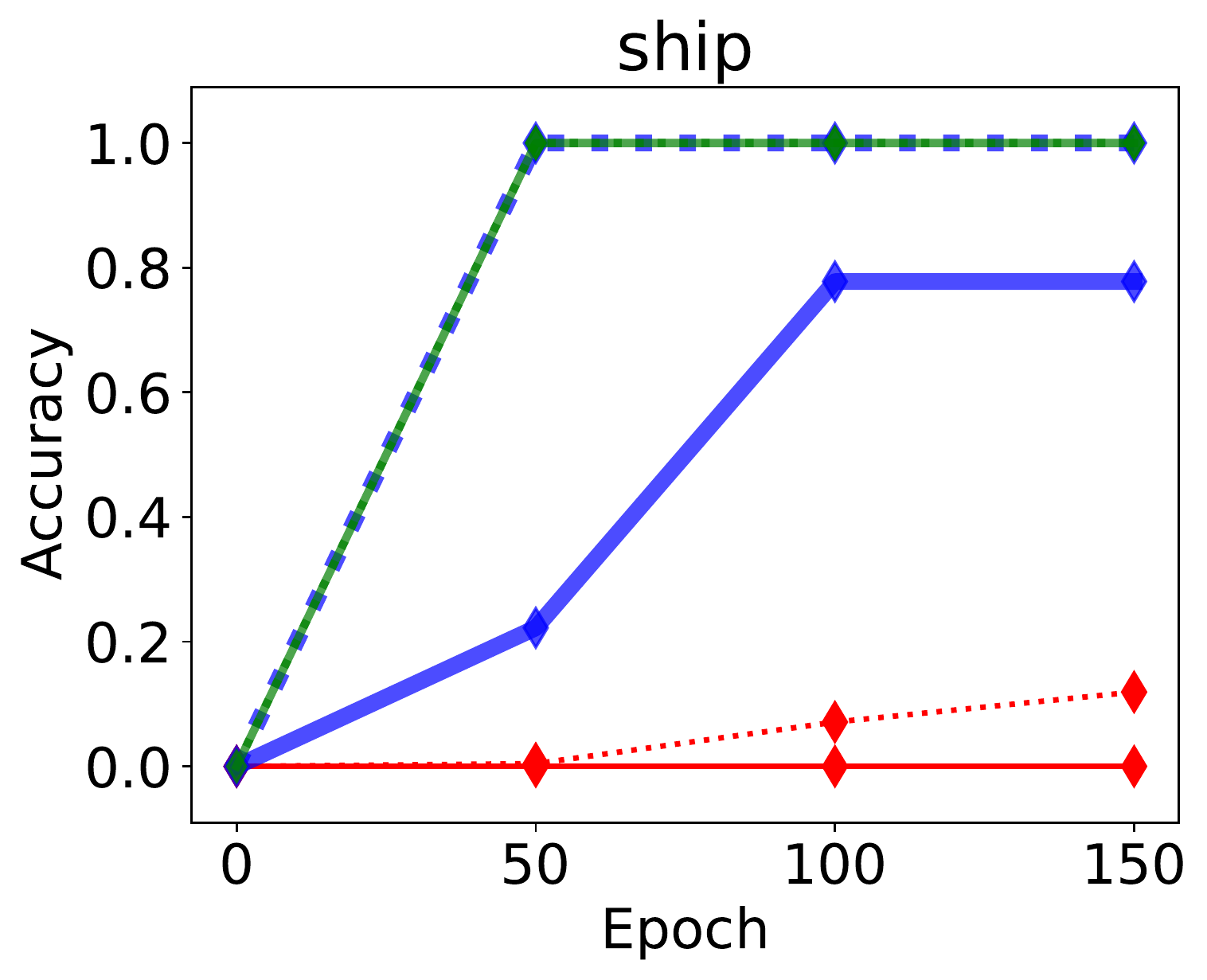}\hfill
    \includegraphics[width=0.1428\textwidth]{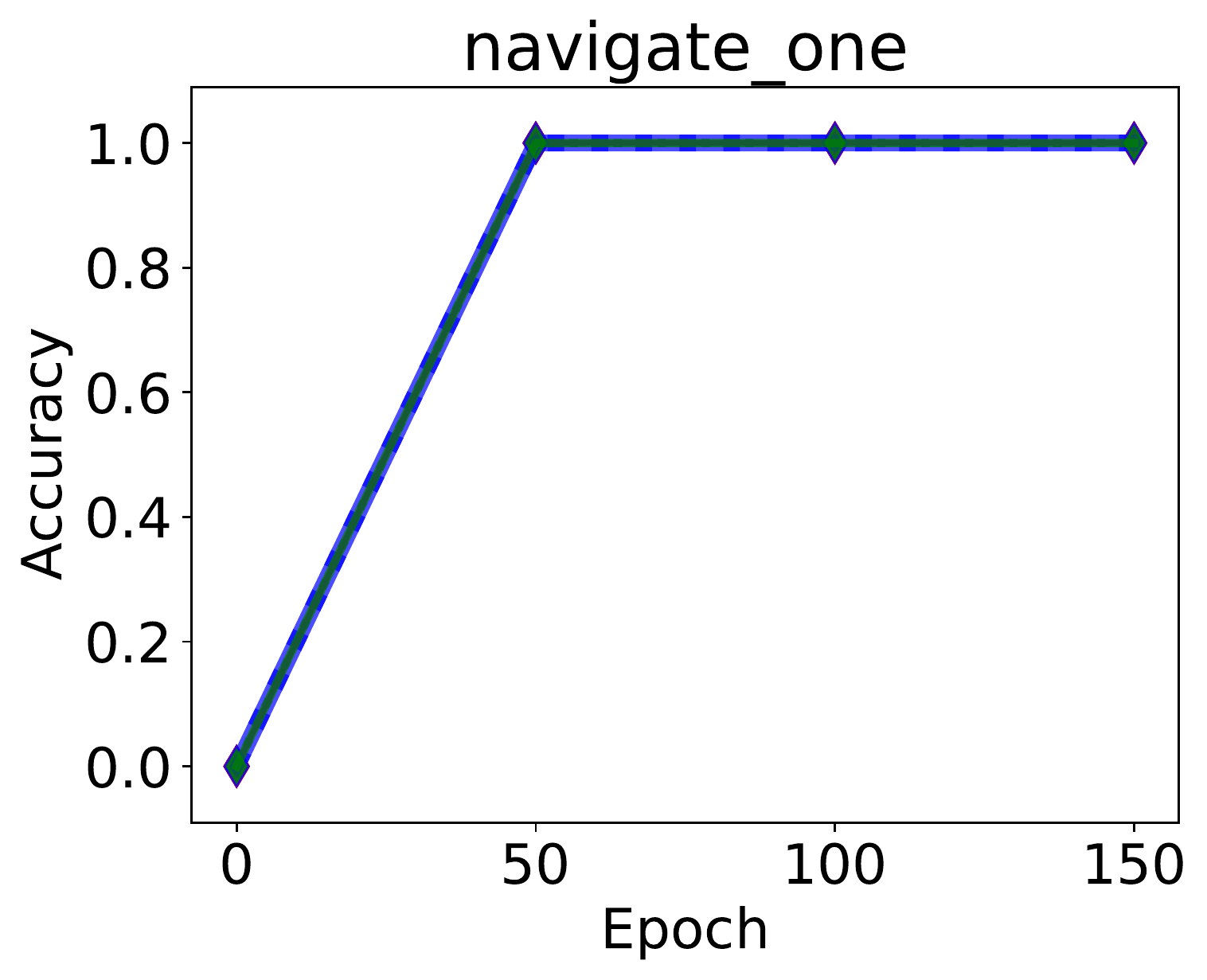}\hfill
    \includegraphics[width=0.1428\textwidth]{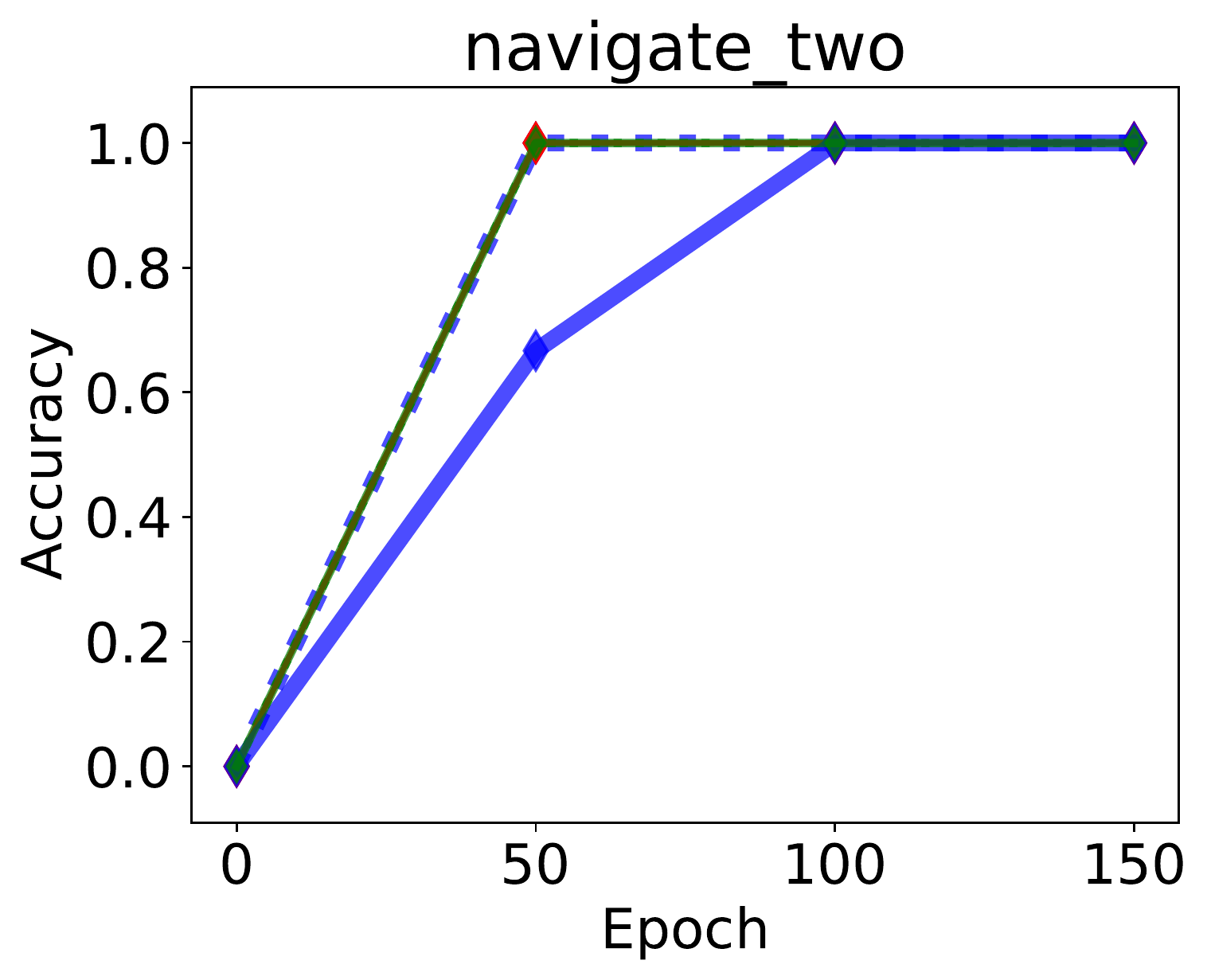}\hfill
    \includegraphics[width=0.1428\textwidth]{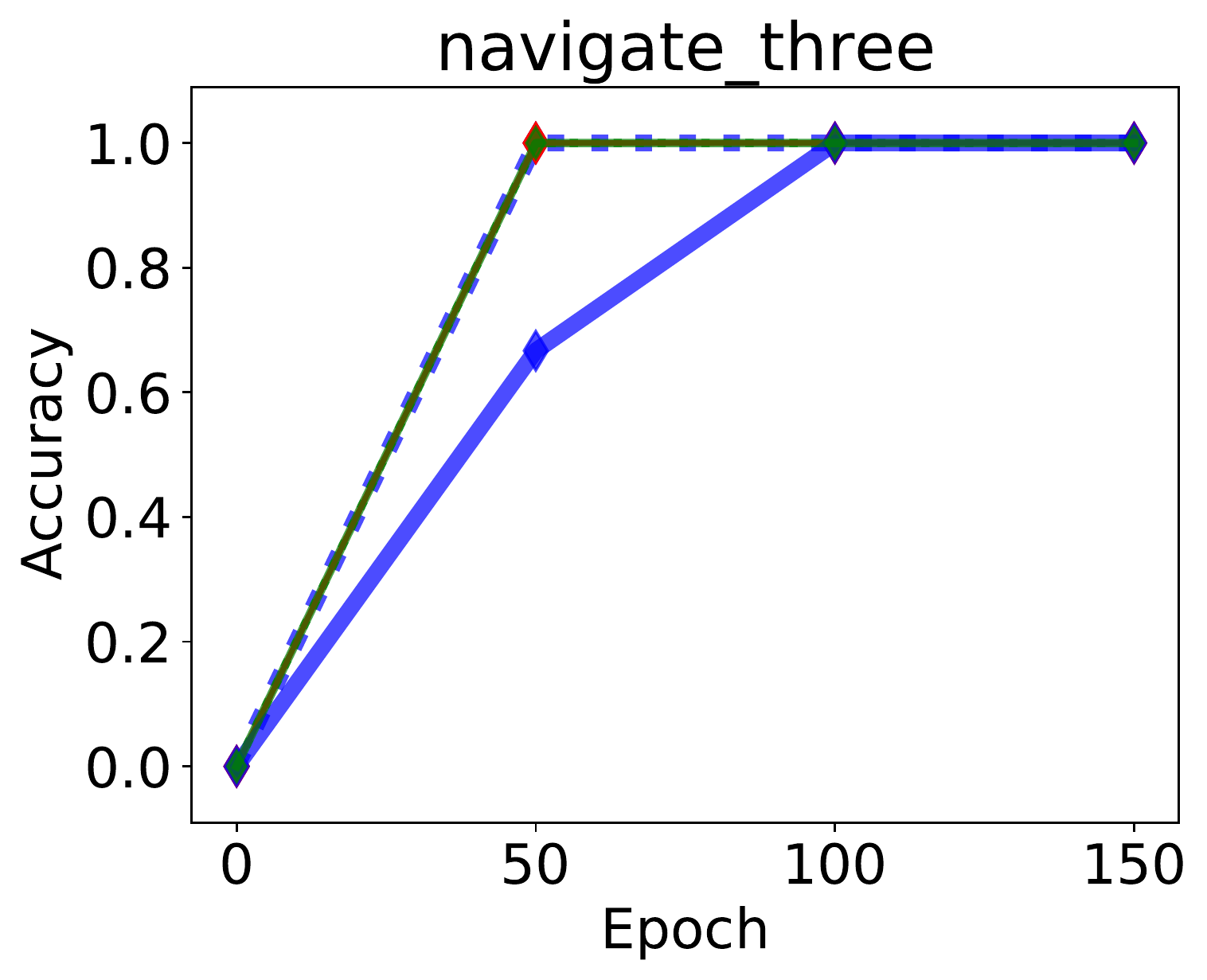}
    \caption{Model accuracy on generating correct LTL task specifications broken down by task class, specified by \textit{seen} vocabulary (model prefix $S$-) and \textit{unseen} vocabulary (model prefix $U$-). Accuracies for tasks from $S$ are plotted with dotted lines. Accuracies for tasks from $U$ are plotted with solid lines.}
    \label{task-class-accuracy}
\end{figure*}
\subsection{Training Pipelines}In our experiments, we refer to models that output contextual queries as ``CQ-CopyNet'' and ``CQ-CopyNet+POS'', and models that output grounded LTL as ``LTL-CopyNet''.

\textbf{CQ-CopyNet + LTL-CopyNet} We jointly trained two CopyNet models, one translating from natural language to contextual queries; the other translating from natural language to LTL. For each domain, we created ``seen'' and ``unseen'' datasets of examples $S$ and $U$, generated from templates populated with objects, attributes, and locations from vocabularies $V_S$ and $V_U$. The models were jointly trained on $T\subset S$, and evaluated separately for accuracy on $S$ and $U$. Only positive examples from joint model evaluations on $S$ were used to derive templated LTL via the CQ-LTL algorithm. Table \ref{dataset-details} shows the manipulation and navigation task class distributions in $S$ and $U$.

\textbf{CQ-CopyNet+POS + LTL-CopyNet} For evaluation, we used part-of-speech (POS) tagging with spaCy along with CopyNet to simultaneously extract noun, adjective, and proper noun contextual query parameters, and to perform task classification of the input natural language task specification. We assessed accuracy overall and across manipulation and navigation task classes in $S$ and $U$ to assess domain transfer and generalization over seen and unseen references to objects.

\subsection{Learning Templated LTL via CQ-LTL}
In order to generate the inputs for the CQ-LTL algorithm, we jointly evaluated the CQ-CopyNet with LTL-CopyNet models, as well as the CQ-CopyNet+POS with LTL-CopyNet models to obtain pairs of contextual query and LTL task specifications. Only jointly accurate translations of the natural language task specifications were used to derive templated LTL for task classes. The set of jointly accurate translations was grouped by class to obtain task class accuracy and confirm presence of positive examples in each task class. Natural language examples with unique contextual query parameters and corresponding LTL were identified in each task class to pass on to the CQ-LTL algorithm to obtain the templated LTL.

The LTL task specifications were transformed into their syntax tree representations, leaf nodes corresponding to the atomic propositions, and non-leaf nodes corresponding to Boolean or temporal logic operators. Each atomic proposition was substituted according to the propositional function and possible objects used to generate the atomic proposition. Next, the association between the contextual query parameters and the propositional function parameters was mapped by parameter matching. The resulting templated LTL for each task class was saved for evaluation on the same $S$ and $U$ datasets as the LTL-CopyNet model.

\subsection{Evaluation}
We refer to our templated LTL models as ``CQ-LTL-CopyNet'' and ``CQ-LTL-CopyNet+POS'', depending on whether POS was enabled. We evaluated the ability for CQ-LTL-CopyNet and CQ-LTL-CopyNet+POS to generate correct instances of grounded LTL on navigation and manipulation natural language queries in $S$ and $U$ by using the contextual query outputs from CQ-CopyNet and CQ-CopyNet+POS. We compared the LTL output by our models to the outputs from LTL-CopyNet.

\section{RESULTS}
Fig. \ref{overall-accuracy} shows overall accuracy of the models on the $S$ and $U$ datasets for manipulation and navigation tasks. Fig. \ref{task-class-accuracy} shows the accuracy for generating correct grounded LTL across task classes specified in $S$ and $U$. For navigation tasks, where atomic propositions were represented directly by natural language vocabulary, LTL-CopyNet was able to generalize successfully due to the copying mechanism of CopyNet. Our templated LTL models were also able to successfully generate the correct grounded LTL. In the case of manipulation tasks, the results confirm our hypothesis that LTL-CopyNet is unable to generate out-of-vocabulary words to use as atomic propositions in the output LTL. This is especially apparent on the unseen manipulation dataset $U$. In comparison, our CQ-LTL-CopyNet and CQ-LTL-CopyNet+POS models were able to achieve above 60\% accuracy on the manipulation tasks from $U$.

The task class breakdown in Fig. \ref{task-class-accuracy} shows which task classes the models had difficulty generating correct grounded LTL. On the seen dataset, the \textit{ship} manipulation task was the source of the majority of the failures, despite comprising roughly 30\% of $S$ at 1350 out of 4650 examples. In the unseen dataset, our templated LTL models primarily failed on the \textit{pickup colored} tasks, which can be attributed to the failure cases in the CQ-CopyNet and CQ-CopyNet+POS models.

Finally, the output grounded LTL was forwarded to a value iteration planning module in the simulated \textit{Toys} environment. Sample planning times for which the output grounded LTL was correct are shown in Table \ref{planning} in order to demonstrate the full pipeline.
\begin{table}[]
    \caption{Sample Task Planning Times with Value Iteration on Toys}
    \label{planning}
    \centering
    \begin{tabular}{|l|c|c|c|}\hline
    Natural Language Task & CQ & Planning & Toy Size \\\hline
    Put the cylinder in the box, & ship & 6m3.72s & 4x1\\
    then put the box in the bedroom & & & \\\hline
    Pickup the sphere & pickup & 1m41.185s& 4x1 \\\hline
    \end{tabular}
\end{table}

\section{DISCUSSION}
We investigated cases in which the trained models failed to produce to the correct output. Our observed failure cases resulted from generating incorrect LTL atomic propositions or from generating incorrect intermediate contextual query inputs. We did not observe any failures attributed to incorrect LTL structure or incorrect contextual query structure. In fact, both CQ-CopyNet and LTL-CopyNet always output the correct structures for contextual queries and LTL during evaluation after 150 epochs of training.

In considering the generation of incorrect atomic propositions, the manipulation and navigation task specifications were expressed with different types of atomic propositions. For navigation, the atomic propositions were drawn solely from natural language references to objects and locations. Thus, as long as the models could predict the correct natural language references to use for the atomic propositions, the output LTL would be correct. However, for manipulation tasks, the atomic propositions were generated as combinations of natural language references to objects and their attributes. The LTL-CopyNet model would not be able to generate atomic propositions lying outside of the vocabulary or input sentences. 

CQ-LTL-CopyNet and CQ-LTL-CopyNet+POS suffered from the failure modes of (1) incorrect contextual query parameters, and (2) language grounding errors. Incorrectly supplied contextual query parameters stem from the performance of CQ-CopyNet and from the spaCy POS tagger. While CQ-CopyNet always returned the correct task classification, it sometimes failed to return the correct parameters, especially on unseen data, indicating an inaccuracy in the copying mechanism. These incorrect parameters were alleviated by POS tagging. However, the spaCy POS tagging models sometimes misclassified verbs, nouns, and adjectives, resulting in the failure cases where an incorrect number of contextual query parameters was specified. For instance, sometimes ``pickup'' was misclassified as a noun when it was used as a verb. Finally, even when correct parameters were supplied, language grounding errors could still occur. In particular, the seen dataset contained the homonyms ``orange'' and ``bag'', where ``orange'' was simultaneously both a color and a fruit, and ``bag'' was simultaneously a container for objects as well as a receptacle for other containers. If these terms were provided, the ambiguous groundings could result in incompletely populated grounded LTL. This issue could be ameliorated by having the grounding function also consider part-of-speech as part of the grounding process.

\section{CONCLUSION}
We have introduced an intermediate task specification representation combining contextual queries with templated LTL, and provided an algorithm for associating a contextual query with templated LTL from single positive examples. Our experiments in a simulated OO-MDP environment indicate that mapping from natural language to our intermediate representation yields improved generalization over novel objects compared to state-of-the-art NL to LTL Seq2Seq models for cases where the atomic propositions cannot be expressed by existing natural language vocabulary. We also discuss errors common in generating correct grounded LTL. Finally, we provided example planning demonstrations with pipeline on example manipulation and navigation tasks.
 
\small
\bibliographystyle{IEEEtran}
\bibliography{sample}

\end{document}